\documentclass[10pt]{article}
\usepackage[preprint]{tmlr}

\usepackage{amsmath,amsfonts,bm}

\def\eqref#1{equation~\ref{#1}}

\def\1{\bm{1}}

\DeclareMathAlphabet{\mathsfit}{\encodingdefault}{\sfdefault}{m}{sl}
\SetMathAlphabet{\mathsfit}{bold}{\encodingdefault}{\sfdefault}{bx}{n}

\usepackage{hyperref}
\usepackage{url}
\usepackage{booktabs}
\usepackage{graphicx}
\usepackage{amsmath}
\usepackage{tikz}
\usetikzlibrary{arrows.meta}

\title{Anisotropy Decides Cosine vs.\ Rank Metrics for Text Embeddings}

\author{\name V.~S.~Raghu Parupudi \email pvsrrkishore@gmail.com \\
      \addr University of California, San Diego}

\def\month{MM}
\def\year{YYYY}
\def\openreview{\url{https://openreview.net/forum?id=XXXX}}

\begin{document}

\maketitle

\begin{abstract}
The standard way to compare two text embeddings is cosine similarity. Scattered
studies report that a different metric does better, but they never pin down the
geometric condition that decides when, or the reason why. We settle both with a
comprehensive empirical study: nineteen parameter-free similarity metrics on
nineteen encoders, from compact sentence transformers up to seven-billion-parameter
large language models, across seven datasets. The answer is geometric. When an
encoder spreads its variance evenly across directions, cosine similarity is the
best parameter-free choice and no other metric helps by a usable margin. When the
variance concentrates into a few dominant directions, a property known as
anisotropy, rank-based and $L_1$-type metrics beat cosine by a clear margin. The
absolute gain is modest, but because cosine starts low on these encoders it is a
sizable relative improvement, around twenty percent on average and largest where
cosine is weakest. What
decides this is the geometry of the embedding space, not how the model was
trained. When the two point different ways, the metric follows the geometry. One
number captures the concentration, namely the fraction of variance held by the
single most dominant dimension, and it predicts how much the alternatives help
across all nineteen encoders, with a rank correlation of $0.86$ and a linear
correlation of $0.95$. To test concentration as the cause
rather than a correlate, we project out the dominant directions: cosine recovers
and the advantage of the other metrics nearly vanishes, but only on the encoders
that were anisotropic to begin with. The effect is directional, not magnitude
based, since it survives normalizing every vector to unit length, and the one
metric that relies on length is the worst of the set. Among parameter-free
metrics, then, cosine is the right tool wherever an encoder is well spread, which
includes the fine-tuned embedders commonly deployed for retrieval and similarity,
and we give a one-number diagnostic for when it is not.
\end{abstract}

\section{Introduction}

Text embeddings turn sentences into vectors, with the idea that similar
sentences end up as nearby vectors. Using them requires a way to measure how
near two vectors are. Search, retrieval, clustering, and paraphrase detection
all come down to this one measurement, and the usual choice is cosine
similarity \citep{reimers2019sentence}.

People have questioned whether cosine is the right default. One line of work
shows that embeddings from language models are anisotropic, which means the
vectors bunch up in a narrow part of the space instead of spreading out
\citep{ethayarajh2019contextual, gao2021simcse}. When the vectors are all
crowded together, even unrelated sentences look similar to cosine, so it loses
its edge. Other papers suggest swapping cosine for a different simple metric.
Some find that Euclidean distance retrieves better in certain settings
\citep{tessari2025dimension}. Others reach for set-overlap measures like the
Tanimoto coefficient from chemistry \citep{bajusz2015tanimoto}. Each of these
comes from a narrow setup, and together they do not add up to a clear rule. No
one has said when a non-cosine metric actually helps, or what about the encoder
decides it.

That is the gap we fill. We do not propose a new metric. We ask a plain
question instead. For a given encoder and task, can any simple metric beat
cosine, and if so, what controls how much it helps? To answer it we vary the two
things that matter. One is the encoder, and we use nineteen of them, from small
sentence transformers up to seven-billion-parameter language models, including
multilingual ones. The other is the metric, and we use nineteen parameter-free
ones taken from the literature on comparing high-dimensional vectors. We test
every encoder and metric on seven datasets covering semantic similarity,
paraphrase detection, and inference.

We treat the rest of the paper as a small investigation. We start with what we
see, ask why it happens, list the explanations that could account for it, and
knock them out one by one until the real one is left. What we see is a clean
split. On some encoders cosine is hard to beat, and the so-called alternatives
turn out to be cosine in disguise. On the others, rank-based and $L_1$-type
metrics beat cosine by about $0.05$ in Spearman correlation, which is a real
margin on these tasks.

The split lines up almost perfectly with how the encoder was trained. Models
trained with a cosine objective sit on the cosine-wins side, and plain language
models sit on the other side. It would be easy to stop there and say training is
the answer. The reason we do not is that training is not what the metric responds
to. The metric responds to the shape of the embedding space, and training is just
the usual way an encoder ends up with one shape or the other. Two encoders make
this clear, because in them shape and training disagree. E5-Mistral is trained
with a cosine objective, yet its space is crowded, and the alternatives help it.
Multilingual BERT is a plain language model, yet its space is well spread, and
the alternatives do not help it. Both follow their geometry, not their training
label.

So we investigate, looking for some property of the encoder, the task, or the
metric that decides the outcome, and we rule out the easy explanations first.
Size is not it, since two models of the same size land on opposite sides. Pooling
is not it, since models that pool the same way spread across the whole range. The
task is not it either, because within every task type the split still falls along
geometry, with crowded encoders gaining and well-spread ones not. Weak cosine is
not it, which we show by holding cosine's own quality fixed and watching the
effect survive. Vector length is not it, since the result barely moves when we
rescale every vector to length one. What is left is how the variance is spread
across directions. The encoders where alternatives help are the ones that pack
most of their variance into a few directions, and a single measure of that
packing predicts the gain across all nineteen encoders with a rank correlation of
$0.86$ and a linear correlation of $0.95$.

A correlation is not proof, so we test the cause by removing it. If those few
crowded directions are what break cosine, taking them out should let cosine
recover and remove any reason for the other metrics. That is what happens. We
project the top directions out, cosine recovers most of what it lost on the
crowded encoders, the advantage of the alternatives nearly vanishes, and on
encoders that were never crowded the same step changes little. This is not a clean
causal proof, since removing the top directions also removes whatever else lives
in them, but the result is what we would expect if the crowded directions are the
cause.

The contribution is the diagnosis, not a metric. The choice of similarity
function matters only when an embedding space is packed into a few directions,
that packing is what training does or fails to do, and one number measures it. The
takeaway is reassuring for practice. Most fine-tuned embedders we tested, the kind
commonly deployed for similarity, are trained to spread their vectors out, so
among parameter-free metrics cosine is already the right call. What the result
adds is a way to tell when cosine is at risk and a simple number to check it.

\section{Related Work}

\paragraph{Anisotropy in embedding spaces.}
Embeddings from language models do not spread out evenly.
\citet{ethayarajh2019contextual} measured that vectors from BERT and GPT-2 sit in
a narrow cone, so two random sentences already look similar under cosine.
\citet{gao2021simcse} tied this to weak similarity performance and showed that a
contrastive objective spreads the vectors out. \citet{mu2018allbutthetop} found
that a few top directions dominate word embeddings and removed them, a trick they
called all-but-the-top. \citet{rudman2022isoscore} gave a score for how evenly a
cloud fills the space, IsoScore. \citet{cai2021isotropy} complicated the
narrow-cone picture, showing the space breaks into clusters and manifolds that are
locally more isotropic than they appear. We use these tools to measure an
encoder's geometry, but we do not try to fix the space. We ask how its shape
decides which metric to use, and we pin down the geometric quantity that controls
the answer.

\paragraph{Rogue and outlier dimensions.}
A closely related line traces anisotropy to a handful of dimensions with oversized
values. \citet{kovaleva2021bert} found outlier dimensions in BERT whose removal
hurts the model, and \citet{puccetti2022outlier} linked them to token frequency.
\citet{sun2024massive} described the same phenomenon in large language models as
massive activations, with single dimensions reaching magnitudes in the thousands.
Closest to our work, \citet{timkey2021bark} named these rogue dimensions and
showed they dominate cosine similarity and hide representational quality, fixing
it by standardizing the dimensions to repair cosine. We instead ask which metric
to use given that these dimensions exist, measure how much they dominate with a
single number, and show by removing them that they drive the metric gap across
nineteen encoders.

\paragraph{Questioning cosine.}
Several papers have asked whether cosine should be the default.
\citet{zhou2022problems} showed it is unreliable for very frequent words.
\citet{steck2024cosine} argued that for some linear models cosine can be arbitrary
and not even unique. \citet{tessari2025dimension} reported that Euclidean distance
can beat cosine for retrieval in high dimensions. \citet{parupudi2025magnitude}
introduced magnitude-aware metrics, an overlap similarity and a hyperbolic-tangent
similarity, and reported that they improve on cosine in mean squared error on some
semantic tasks. We include both metrics here and find that their advantage is
about direction, not magnitude. It survives length normalization, and the
calibration gap closes under a single fitted calibrator, so the improvement is not
a property of magnitude. A separate tradition in chemistry uses the Tanimoto
coefficient, the extended Jaccard measure, to compare real-valued vectors
\citep{tan2005introduction, bajusz2015tanimoto}. These results hint that the
metric matters, but they come from different settings and never say when each one
wins. We put all of them into one comparison and find that several are just cosine
in another form, while the ones that really differ help only under a specific
condition on the geometry.

\paragraph{Fixing the space instead of the metric.}
A parallel line keeps cosine and repairs the space. BERT-flow maps the vectors to
a more even distribution with a normalizing flow \citep{li2020sentence}. Whitening
transforms them so their covariance is the identity \citep{su2021whitening,
huang2021whiteningbert}. \citet{haemmerl2023anisotropy} studied anisotropy and
outliers in multilingual models and how removing them affects cross-lingual
matching. These methods change the embeddings and then use cosine. We instead hold
the embeddings fixed and study the metric, which separates the role of the space
from the role of the comparison. The two views meet in our removal test, where
taking out the crowded directions is a minimal version of these repairs.

\paragraph{Distances in high dimensions.}
\citet{aggarwal2001surprising} argued that in high dimensions fractional $L_p$
norms and $L_1$ distance are often more meaningful than $L_2$, because $L_2$ lets
a few large coordinates take over. Our findings tie this older result to modern
embeddings: the metrics that beat cosine on the crowded encoders are exactly the
rank-based and $L_1$-type ones that do not let a few coordinates take over.

\paragraph{Encoders and benchmarks.}
Sentence transformers fine-tune a language model with a contrastive objective so
cosine lines up with meaning \citep{reimers2019sentence}. Newer models such as E5
\citep{wang2022e5} and BGE \citep{xiao2023cpack} use the same recipe at scale, and
LLM2Vec turns decoder-only models into embedders \citep{behnamghader2024llm2vec}.
The MTEB benchmark gathers many of these models and tasks in one place
\citep{muennighoff2023mteb}. We draw both the contrastively trained embedders and
the plain base models from this landscape, putting encoders on both sides of the
cosine objective.

\section{Experimental Setup}

We want to know when a simple metric beats cosine and what predicts it. So we
vary three things: the encoder that makes the embeddings, the metric that
compares them, and the dataset that says how similar the pairs really are. We
go through each one, then the geometry measures and how we score everything.

\subsection{Encoders}
We use nineteen encoders, picked to spread along the axis we think matters,
which is whether the model was trained with a cosine objective. On one side are
the contrastively trained embedders \citep{reimers2019sentence}, which fine-tune
a base language model with a contrastive objective: all-MiniLM-L6 and
all-MiniLM-L12 \citep{wang2020minilm}, all-mpnet-base and paraphrase-mpnet-base
\citep{song2020mpnet}, BGE-base and BGE-large \citep{xiao2023cpack}, E5-large and
multilingual-E5-large \citep{wang2022e5},
E5-Mistral-7B \citep{wang2024e5mistral}, and SFR-Embedding-Mistral
\citep{meng2024sfr}. These are built
to be compared with cosine, and several of them already normalize their output
to length one. On the other side are models that were not trained this way:
BERT \citep{devlin2019bert}, RoBERTa \citep{liu2019roberta}, ELECTRA
\citep{clark2020electra}, multilingual BERT \citep{devlin2019bert}, GPT-2
\citep{radford2019gpt2},
Pythia-410M \citep{biderman2023pythia}, Qwen2.5-1.5B and Qwen2.5-7B
\citep{qwen2024report}, and Mistral-7B \citep{jiang2023mistral}. For these we
either average the token vectors or take the last token, with no extra
fine-tuning. The set runs from twenty-two
million up to seven billion parameters and covers both English and multilingual
models. Table~\ref{tab:encoders} lists each encoder with its size, pooling, and
main geometry numbers.

\subsection{Similarity metrics}
We compare nineteen parameter-free metrics from the literature on comparing
high-dimensional vectors, drawing on standard catalogs of distance and
similarity measures \citep{cha2007comprehensive, levy2024guide}. They fall into
a few groups. The angular and inner-product group has cosine, the raw dot
product, angular similarity, correlation similarity, and several normalized inner
products: the Tanimoto coefficient \citep{bajusz2015tanimoto}, the Dice
coefficient \citep{dice1945measures}, an overlap similarity, and a
hyperbolic-tangent version. The Minkowski group has $L_1$ (Manhattan), $L_2$
(Euclidean), $L_\infty$ (Chebyshev), a fractional $L_p$ distance with $p=0.5$
\citep{aggarwal2001surprising}, Canberra distance \citep{lance1966computer}, and
Bray-Curtis similarity \citep{bray1957ordination}. We also include a Gaussian
kernel, three information-geometry measures that turn each vector into a
distribution and compare them with the Jensen-Shannon \citep{lin1991divergence},
Hellinger \citep{hellinger1909neue}, and Bhattacharyya \citep{bhattacharyya1943measure}
divergences, and a rank-based measure built on Spearman rank correlation
\citep{spearman1904proof} that compares the rank order of the two vectors. That
is nineteen in all. None of these is ours. They are a broad sample, so that
whatever pattern we find comes from the data and not from a metric we favor.

\subsection{Datasets}
We test on seven datasets in three groups. Semantic similarity uses STS-B from
the GLUE benchmark \citep{wang2019glue}, SICK-R \citep{marelli2014sick}, and
STS16 \citep{cer2017semeval}, which give a graded similarity score. Paraphrase
detection uses Quora and PAWS \citep{zhang2019paws}, which give a yes-or-no
duplicate label. Inference uses SNLI \citep{bowman2015snli} and MultiNLI
\citep{williams2018multinli}, which we turn into a similarity task by treating
entailment as a one and everything else as a zero. We scale all the gold scores
to lie between zero and one. The datasets and their splits are standard, and we
give the exact sources in the appendix.

\subsection{Geometry measures}
For each encoder we describe the shape of its embedding cloud with a few numbers,
computed on the pooled embeddings of all the datasets. IsoScore says how evenly
the cloud fills the space \citep{rudman2022isoscore}. The mean cosine between
random pairs and the mean cosine to the average direction both say how much the
vectors share a common heading. The top-one and top-ten variance ratios say how
much of the spread sits in the leading directions. The participation ratio says
roughly how many dimensions the cloud really uses. We also report one number we
call rogue-dimension dominance, which is the share of the total variance held by
the single biggest-variance dimension. That last number turns out to be the one
that predicts our results.

\subsection{Scoring}
For every encoder, dataset, and metric, we score each sentence pair and compare
those scores to the gold labels. We report Spearman rank correlation for ranking
quality on all the datasets, and ROC-AUC and PR-AUC on the four datasets with
yes-or-no labels. For calibration we report mean squared error after mapping the
scores into the unit range, expected calibration error, and mean squared error
after fitting a single isotonic calibrator on held-out data. We run every metric
two ways. The raw way uses the embeddings as the encoder produces them. The
normalized way first scales every vector to length one, which lets us separate
the role of vector length from the role of direction. To keep memory in check
on the largest encoders, we score in chunks of at most two thousand pairs and
stitch the full score vectors back together before computing any statistic, so
the chunking changes none of the reported numbers.

For significance we use distribution-free tests, since the gains are not normal
and the encoders are few. We test whether a metric beats cosine with a Wilcoxon
signed-rank test over the encoder-dataset cells, and whether one group of
encoders gains more than another with a Mann-Whitney test. For the
removal experiment we use a paired Wilcoxon test over encoders, comparing each
encoder to itself before and after removal. For the correlation between geometry
and gain we report the $p$-value of the correlation along with a bootstrap
interval and a leave-one-out check. We treat $p$ below $0.05$ as significant.

\section{Results}

We lay the results out as an investigation. Section~\ref{sec:obs} states what we
see. Section~\ref{sec:why} lists the things that could cause it and knocks out
the ones that fail. Section~\ref{sec:cause} pins down the real cause and tests
it by removing it. Section~\ref{sec:mech} explains why it works and what it means
in practice. Throughout, we split the encoders by the shape of their space rather
than by how they were trained. We call an encoder crowded, or anisotropic, when
its rogue-dimension dominance is above $0.01$, and well spread, or isotropic, when
it is below. Nine encoders are crowded and ten are well spread. This split tracks
training closely but not exactly. E5-Mistral was trained with a cosine objective
yet is crowded, and multilingual BERT is a plain language model yet is well
spread, so the two groups are not the same as base versus contrastively trained.

\subsection{What we see: cosine wins, except on crowded encoders}
\label{sec:obs}

The first thing we see is that cosine is almost impossible to beat on the
well-spread encoders. Across the ten of them, the best alternative metric beats
cosine by only $0.001$ in Spearman on average, and it gains more than $0.01$ in
just three percent of the encoder-dataset cells. This tiny gain is not
significant. A one-sided Wilcoxon signed-rank test over the seventy well-spread
cells cannot reject the idea that the alternatives fail to beat cosine
($p = 0.54$). On these encoders, the choice of metric does not matter.

One reason is that a lot of the proposed alternatives are really cosine in
another form. Cosine, angular similarity, and correlation similarity rank pairs
in almost exactly the same order, differing by at most $0.0002$ in Spearman across
every encoder and dataset. When the vectors have length one, these functions are
order-preserving rescalings of each other, so they cannot put the pairs in a
different order. A few other normalized inner products, such as the Tanimoto
coefficient and a hyperbolic-tangent version, agree with cosine on the
well-spread encoders but drift from it on the crowded ones, where they end up a
little worse than cosine. Calling any of these a real alternative to cosine is a
mistake, and the data shows it plainly.

\begin{figure}[t]
\centering
\includegraphics[width=0.6\linewidth]{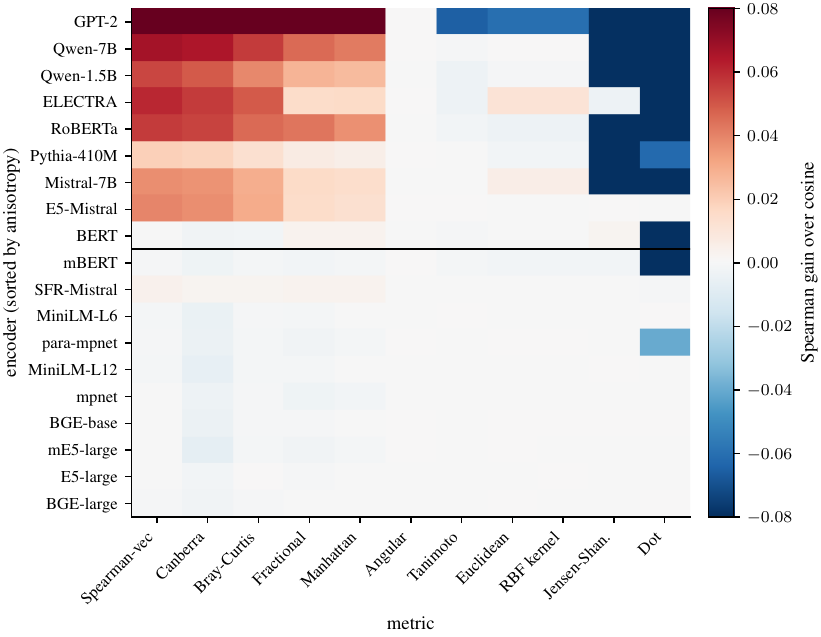}
\caption{Spearman gain over cosine for every metric and encoder. Encoders are
sorted by anisotropy, and the black line separates the nine crowded encoders
above from the ten well-spread ones below. The rank-based and $L_1$-type metrics
(red, upper-left block) gain only on crowded encoders. The lower block is nearly
white, since on well-spread encoders no metric beats cosine. The dot product and
the information-geometry metrics lose, shown in blue.}
\label{fig:heatmap}
\end{figure}

The second thing we see is the reverse. Across the nine crowded encoders, the
best alternative beats cosine by $0.055$ in Spearman on average, and it does so in
eighty-seven percent of cells. This gain is highly significant. Over the
sixty-three crowded cells the same Wilcoxon test puts the gain well above zero
($p = 3 \times 10^{-12}$), and a Mann-Whitney test confirms the crowded gain is
larger than the well-spread one ($p = 1 \times 10^{-21}$). The winners are always
the same kind. The rank-based metric gains $0.053$, Canberra gains $0.051$,
Bray-Curtis gains $0.039$, the fractional $L_p$ distance gains $0.034$, and
Manhattan gains $0.028$. Each of these five is significant on its own across the
crowded cells, with every $p$ below $10^{-10}$.
These are the rank-based and $L_1$-type metrics, and they win on exactly the
encoders where cosine has trouble. The gain is not an artifact of picking the best
of several metrics, since the single rank-based metric on its own gains $0.053$,
almost the whole margin. Table~\ref{tab:bymetric} gives the full per-metric
ranking for both groups, and Figure~\ref{fig:heatmap} shows the same split as a
grid of gains over every encoder and metric.

\begin{figure}[t]
\centering
\includegraphics[width=0.4\linewidth]{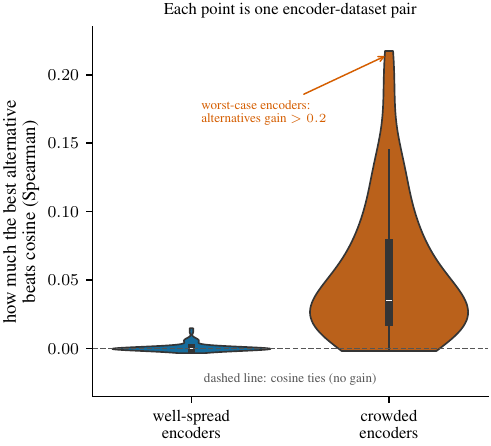}
\caption{How much the best alternative metric beats cosine, one point per
encoder-dataset pair, split by encoder geometry. A crowded encoder is one whose
variance piles into a few directions, and a well-spread one is the opposite. On
well-spread encoders the gain sits at zero, so cosine cannot be beaten. On crowded
encoders the gain spreads upward and reaches past $0.2$ in the worst cases. The
choice of metric only matters for the crowded group.}
\label{fig:violin}
\end{figure}

So cosine is the best choice on the well-spread encoders and is clearly beaten on
the crowded ones. Figure~\ref{fig:violin} shows how far apart the two groups are.
The rest of the section asks what about an encoder decides this.

\subsection{Why? Ruling out the easy answers}
\label{sec:why}

Several things could explain the split: model size, the pooling method, the task,
how good cosine was to begin with, vector length, and the shape of the embedding
space. We test all but the last here, and none of them holds up.

\paragraph{Size.}
Maybe bigger models are more prone to the effect, but they are not. The rank
correlation between model size and the metric gain is only $0.46$, and on raw
size it is weaker still. The clear test is to fix the size and look. Among the
seven-billion models, Qwen2.5-7B gains $0.067$ but SFR-Embedding-Mistral gains
$0.004$, a fifteen-fold difference at the same size. Among the hundred-million
models, ELECTRA gains $0.060$ and RoBERTa gains $0.057$, while BGE-base, mpnet,
and paraphrase-mpnet gain nothing. Models of the same size end up on opposite
sides, so size is not the cause. The weak correlation it does show is a side
effect, because the biggest models in our set happen to be the untrained base
ones.

\paragraph{Pooling.}
The base models often take the last token, so pooling could be the hidden cause,
but it is not. The mean-pooled models on their own already cover almost the whole
range, from $-0.001$ for paraphrase-mpnet to $0.148$ for GPT-2. Within that group
the gain still follows the geometry, not the pooling, so pooling is out.

\paragraph{Task.}
Maybe the alternatives only help on certain task types, but the split is the same
inside each one. We group the seven datasets into similarity, paraphrase, and
inference, and within every group the crowded encoders gain while the well-spread
ones do not. The crowded-encoder gain is $0.064$ on similarity, $0.058$ on
inference, and $0.037$ on paraphrase, and the matching well-spread gain is under
$0.001$ in all three. The geometry split holds task by task, so the task is not
the cause.

\paragraph{Cosine being weak.}
Maybe the alternatives only help where cosine was already bad, no matter the
shape of the space. Cosine quality and the gain do correlate at $-0.75$, so this
is worth checking with care. A partial correlation settles most of it. Holding
cosine quality fixed, the link between rogue-dimension dominance and the gain is
still $0.93$ and significant ($p = 1 \times 10^{-8}$), barely down from $0.95$.
Holding the geometry fixed instead, the effect of cosine quality falls to
$-0.55$. The geometry drives the gain on its
own, and the link to cosine quality is mostly there because weak cosine and a
crowded space tend to show up together. We cannot fully separate the two, because
our set has only one encoder that is weak at cosine while still being well spread,
and we return to this in the limitations.

\paragraph{Vector length.}
The obvious guess is that the winning metrics use vector length, which cosine
throws away, but they do not. The gain barely changes when we first scale every
vector to length one. The average best-alternative gain moves only from $0.026$
to $0.026$, a change of less than a thousandth, and the link between geometry and
gain stays at $0.95$ on the normalized vectors. If the advantage came from
length, normalizing would erase it. On top of that, the
one metric that uses length directly, the raw dot product, is the worst of the
whole set. It trails cosine by $0.082$ on raw vectors and only matches cosine once
the vectors are normalized, at which point it is cosine. The effect is about the shape of the
space, not the length of the vectors.

\subsection{The cause: a few crowded directions, confirmed by removing them}
\label{sec:cause}

With the easy answers gone, the shape of the space is what is left, and now we
can say which feature of the shape matters. We measure the crowding with
rogue-dimension dominance, the share of the total variance held by the single
biggest-variance coordinate. This is a basis-dependent number, computed on the
raw coordinates rather than on principal directions, and we report the principal
version too. Across the nineteen encoders, rogue-dimension dominance predicts the
best-alternative gain at a rank correlation of $0.86$ and a linear correlation of
$0.95$, with a bootstrap ninety-five percent interval of $[0.840, 0.990]$ on the
linear value. Both correlations are significant, the rank one at
$p = 3 \times 10^{-6}$ and the linear one at $p = 2 \times 10^{-10}$. We lead with
the rank correlation because the linear one leans on a
single extreme encoder. GPT-2 sits far out from the rest, and dropping it pulls
the linear correlation down to $0.91$. The relationship is not just a gap between
two clusters. Within the nine crowded encoders alone, where there is no cluster
gap to lean on, rogue-dimension dominance still predicts the gain at a linear
$0.95$ and a rank $0.88$. Dropping each encoder in turn keeps the full-set
correlation between $0.909$ and $0.966$. The encoders are also not all
independent, since several share an architecture, so we collapse them into eleven
model families and average within each. The correlation holds at a linear $0.97$
and a rank $0.91$ across the eleven families, so it is not an artifact of counting
near-duplicate encoders as separate points.

\begin{figure}[t]
\centering
\includegraphics[width=\linewidth]{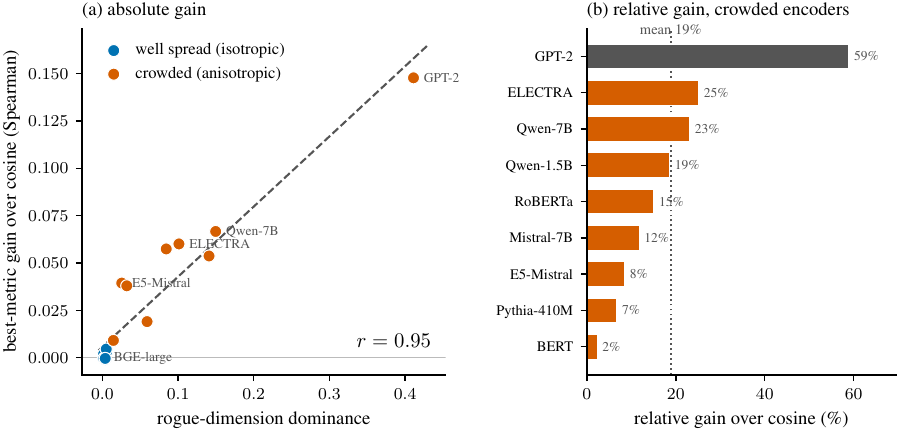}
\caption{(a) The metric gain against rogue-dimension dominance, one point per
encoder. The well-spread encoders cluster at the origin with no gain and no
dominant direction, while the crowded encoders climb the line. A single geometry
number predicts the gain across all nineteen encoders, with a rank correlation of
$0.86$ and a linear correlation of $0.95$. (b) The same gains for the crowded
encoders expressed relative to their cosine baseline, sorted by size. Because
cosine starts low here, the modest absolute gains are sizable in relative terms,
about twenty percent on average (dotted line). GPT-2 (grey) is a clear outlier at
fifty-nine percent, while the rest run from a few percent up to the mid-twenties.}
\label{fig:scatter}
\end{figure}

The top-one principal variance ratio tells the same story, predicting the gain at
$0.940$, so the basis-dependent number and the principal one agree. The broader
measures of anisotropy do worse on the linear scale. IsoScore and the
participation ratio reach about $0.55$, and the mean random-pair cosine reaches
$0.69$. On the rank scale these broader measures come closer, since much of
rogue-dimension dominance's linear edge comes from GPT-2, so we do not claim it is
the uniquely correct measure, only that some measure of variance piled into a few
directions is what predicts the gain. Figure~\ref{fig:scatter} shows the
relationship.

A correlation is still not proof, so we test the cause by taking it away. If a few
crowded directions are what break cosine, then removing them should let cosine
recover and should shrink the advantage of the other metrics. We project out the
top principal directions from each encoder, the all-but-the-top trick
\citep{mu2018allbutthetop}, and score everything again. We sweep the number of
removed directions and the effect grows smoothly with it. On the nine crowded
encoders, removing one direction already cuts the best-alternative advantage by
thirty-eight percent and lifts cosine from a mean Spearman of $0.324$ to $0.364$.
Removing five cuts the advantage by seventy-six percent, and removing ten cuts it
by eighty-seven percent, with cosine reaching $0.417$. Both effects on the crowded
group are significant. A paired Wilcoxon test over the nine encoders shows cosine
rises after removal and the metric advantage falls, each at $p = 0.002$, which is
the smallest value the test can give for nine encoders and means every one moved
the same way. We report the ten-direction result as the endpoint of this sweep,
not as a tuned choice. On the ten
well-spread encoders the same step does the opposite. It lowers cosine a little,
by $0.014$ on average at ten directions, because here the leading directions carry
real signal rather than noise, and the metric advantage stays near zero
throughout. How much cosine recovers tracks how crowded the encoder was. GPT-2,
the worst case, sees its metric advantage fall from $0.148$ to $0.016$, while
BGE-large, which had no crowded directions to take out, just loses a little
accuracy. One worry with this test is that it might be circular. We remove the top
directions, and rogue-dimension dominance is built from the top direction, so
maybe removing any ten directions would help just as much. To check this we run a
control. Instead of the top ten directions we remove ten random orthogonal
directions, averaged over five random draws per encoder, and otherwise repeat the
test. On the crowded encoders random removal barely moves anything. It shrinks the
metric advantage by only ten percent, from $0.055$ to $0.049$, against the
eighty-seven percent that removing the top ten directions erases. The
top-ten advantage after removal is far smaller than the random-ten advantage,
significantly so across the nine encoders ($p = 0.002$, paired Wilcoxon). On
GPT-2 the contrast is sharpest. Removing the top ten directions cuts its advantage
from $0.148$ to $0.016$, while removing ten random directions leaves it at $0.21$,
even a touch higher than before. So the effect is specific to the crowded
directions and is not a generic side effect of dropping ten dimensions.

\begin{figure}[t]
\centering
\includegraphics[width=0.7\linewidth]{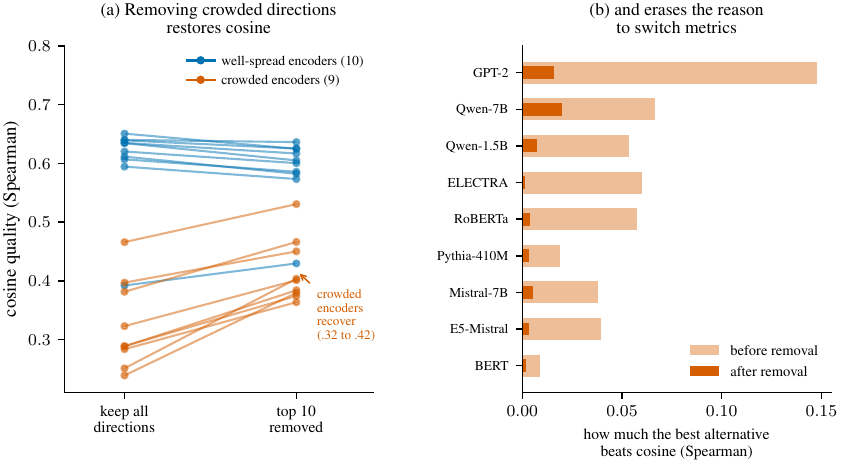}
\caption{The causal test. We project out the top ten principal directions from
each encoder and rescore. Panel (a) tracks cosine quality for each encoder before
and after this removal. On crowded encoders (orange) cosine jumps up, from a mean
of $0.32$ to $0.42$, because the removed directions were mostly noise. On
well-spread encoders (blue) cosine dips a little, because there the removed
directions carried real signal. Panel (b) shows the same removal on the crowded
encoders from the metric side. The bar is how far the best alternative beats
cosine, wide before removal and nearly gone after, so taking out the crowded
directions removes the reason to use anything other than cosine.}
\label{fig:causal}
\end{figure}

We still stop short of calling this a clean causal proof, since removing the top
directions also removes whatever else lives in them, and on the well-spread
encoders we can see that those directions hold signal. But the random-direction
control rules out the most obvious confound, and the whole pattern is what we
would expect if the crowded directions are the cause and is hard to explain if
they are only a bystander. Figure~\ref{fig:causal} shows the split.

\subsection{Mechanism and consequences}
\label{sec:mech}

\begin{figure}[t]
\centering
\includegraphics[width=0.8\linewidth]{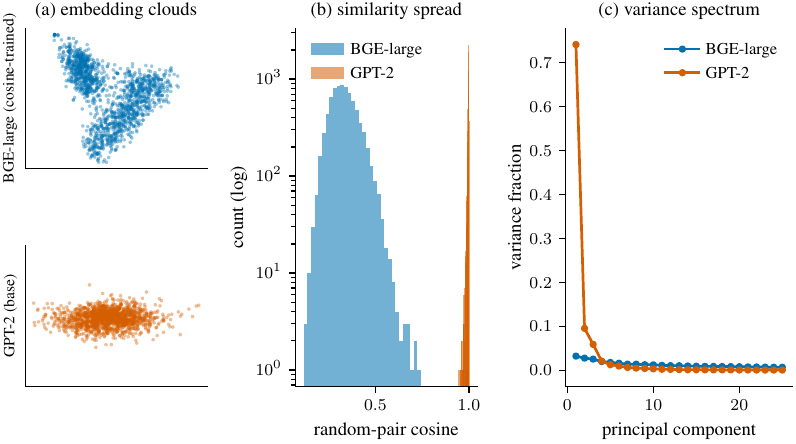}
\caption{The geometry behind the result, for a well-spread encoder (BGE-large)
and a crowded one (GPT-2). Panel (a) projects each embedding cloud onto its
own top two directions. BGE-large fills the plane, while GPT-2 collapses onto a
line. Panel (b) shows the distribution of cosine between random pairs on a log
scale. BGE-large spreads its values, while GPT-2 piles almost every pair near
one. Panel (c) shows the variance spectrum, where GPT-2 places most of its
variance in the first direction.}
\label{fig:hero}
\end{figure}

\begin{figure}[t]
\centering
\begin{tikzpicture}[>=stealth, font=\small]

\begin{scope}[shift={(0,0)}]
  \draw[->] (-1.6,0) -- (1.6,0);
  \draw[->] (0,-1.6) -- (0,1.6);
  \foreach \a/\r in {20/1.0, 75/0.8, 130/1.1, 200/0.9, 260/1.0, 310/0.85,
                     45/0.6, 160/0.65, 240/0.7, 330/0.6, 100/1.15, 290/1.1} {
    \fill[blue!60] (\a:\r) circle (1.3pt);
  }
  \draw[->, thick, blue!70!black] (0,0) -- (35:1.25);
  \draw[->, thick, blue!70!black] (0,0) -- (120:1.15);
  \draw (35:0.55) arc (35:120:0.55);
  \node[align=center] at (0,-2.15) {\textbf{(a) well-spread space}\\[2pt]
    a wide angle separates the\\two example vectors};
\end{scope}

\begin{scope}[shift={(5.4,0)}]
  \draw[->] (-1.6,0) -- (1.6,0) node[right, xshift=1pt] {\scriptsize rogue dim};
  \draw[->] (0,-1.6) -- (0,1.6);
  \foreach \x/\y in {0.9/0.08, 1.1/-0.05, 0.7/0.1, 1.25/0.04, 0.8/-0.09,
                     1.0/0.0, 0.6/0.06, 1.15/-0.03, 0.95/0.07, 1.05/-0.06,
                     0.75/0.03, 1.2/0.05} {
    \fill[orange!75!red] (\x,\y) circle (1.3pt);
  }
  \draw[->, thick, orange!70!black] (0,0) -- (1.25,0.10);
  \draw[->, thick, orange!70!black] (0,0) -- (1.20,-0.08);
  \draw (4:0.6) arc (4:-4:0.6);
  \node[align=center] at (0,-2.15) {\textbf{(b) one rogue direction}\\[2pt]
    both vectors point the same way,\\so cosine cannot tell them apart};
\end{scope}

\end{tikzpicture}
\caption{Why cosine fails under a dominant direction. In a well-spread space (a)
the variance spreads over many directions and the angle between two vectors
reflects their similarity. When one direction holds most of the variance (b),
every vector points mostly along it, the angles between pairs shrink toward zero,
and cosine can no longer tell pairs apart. Rank-based and $L_1$-type metrics do
not normalize by the $L_2$ geometry, so the dominant direction does not capture
them.}
\label{fig:schematic}
\end{figure}
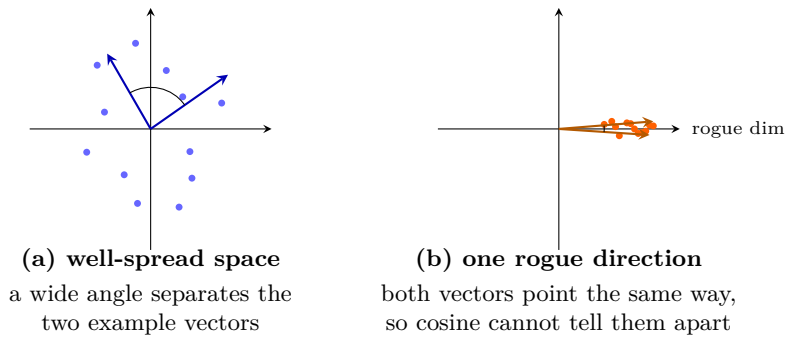

The reason comes straight from what cosine does. Cosine is a normalized inner
product, so a direction that holds a big share of the variance also takes over
the inner product. On a crowded encoder one direction can hold a huge share.
GPT-2 puts forty-one percent of its variance in a single coordinate, really uses
only about two of its seven hundred sixty-eight dimensions, and gives two random
sentences a mean cosine of $0.996$. When one direction is that dominant, nearly
every pair looks alike to cosine, so it cannot tell them apart.
Figure~\ref{fig:hero} shows this geometry, and Figure~\ref{fig:schematic} gives
the intuition. The winning metrics dodge the trap in two ways. The rank-based
metric looks at the order of the coordinates instead of their values, so the one
giant coordinate counts as just the largest rank and nothing more. The $L_1$-type
metrics add up absolute differences rather than squared ones, so a single big
coordinate does not get to dominate the way it does under $L_2$. This is the old
reason that $L_1$ and fractional distances hold up better in high dimensions
\citep{aggarwal2001surprising}. Both kinds of metric recover the signal that
cosine loses to the crowded direction.

\paragraph{The gain is real, not a quirk of rank correlation.}
We double-check the result with ROC-AUC on the four yes-or-no datasets.
Rogue-dimension dominance predicts the AUC gain over cosine at a linear $0.898$,
close to its value for Spearman. On the crowded encoders the AUC gain is $0.028$
on average and significant across cells ($p = 3 \times 10^{-11}$, Wilcoxon).
GPT-2 gains $0.077$ in ROC-AUC, a real improvement in telling duplicate pairs
from non-duplicate ones. So the alternatives improve actual classification, not
just a rank-correlation number.

\paragraph{Calibration is a different question.}
The way metrics differ in calibration has nothing to do with telling pairs
apart. Once we fit a single isotonic calibrator on held-out data, the spread in
mean squared error across metrics shrinks twentyfold, from $0.052$ to $0.0025$.
We measure this spread as the standard deviation of the mean squared error across
metrics within each encoder-dataset cell, averaged over cells on the
length-normalized vectors. The shrinkage happens on every encoder no matter its
shape. Any calibration edge a
metric seems to have is wiped out by one order-preserving rescaling. So
calibration is not a reason to switch metrics, and we keep it apart from the
ranking story.

\paragraph{Earlier claims about other metrics do not hold in general.}
The same sweep lets us check two earlier claims. Euclidean distance does not beat
cosine here. It trails cosine by $0.0002$ on the well-spread encoders and by
$0.006$ on the crowded ones, since with length-one vectors it moves in step with
cosine and on raw vectors it is a touch worse. The information-geometry metrics
are fine on well-spread encoders, where they trail cosine by only $0.0003$, but
they fall apart on crowded ones, trailing by $0.143$, because the dominant
direction swamps the softmax they rely on. Neither claim holds as a general rule.

\paragraph{Where the gain shows up.}
The gain is biggest where cosine is weakest. On the crowded encoders the best
alternative improves similarity tasks by $0.064$, where cosine sits at $0.496$,
improves inference by $0.058$, where cosine sits at only $0.121$, and improves
paraphrase detection by $0.037$, where cosine sits at $0.269$. The worse cosine
does on a task, the more the geometry-robust metrics make up. Because cosine
starts low on the crowded encoders, these absolute gains are sizable in relative
terms. The mean best-alternative gain is a nineteen percent relative improvement
over cosine, with a median near fifteen percent across the nine crowded encoders
and a clear outlier in GPT-2 at fifty-nine percent, and the lowest-cosine task,
inference, gains forty-eight percent. This does not lift these encoders to the
level of a well-spread embedder, whose cosine score is higher to begin with. It
recovers a real part of what the crowded geometry was costing cosine. The same
split holds
if we drop the two inference datasets and look only at the five graded-similarity
and paraphrase ones, where the crowded-encoder gain is still $0.053$, so the
result does not rest on our rough mapping of inference to a similarity score. The
gain is positive on every one of the seven datasets for the crowded encoders, from
$0.030$ on PAWS to $0.087$ on STS-B, and near zero on all seven for the
well-spread encoders, so no single dataset drives the effect.
Table~\ref{tab:bydataset} in the appendix gives the full per-dataset breakdown.

\paragraph{What this means in practice.}
Most fine-tuned embedders we tested, the kind people actually use for similarity,
are trained to spread their vectors out, and on those, among parameter-free
metrics, cosine is the right call. Switching metrics there buys $0.001$ on average, against a cosine Spearman
of $0.602$. The alternatives help on crowded encoders, where cosine sits at
$0.324$, and those are mostly plain language models used without fine-tuning. Such
encoders do show up in practice. Anyone who builds a retrieval or clustering
pipeline on the hidden states of a base or instruction-tuned language model,
which is common when a task has no fine-tuned embedder or when one wants to skip a
training step, is using exactly this kind of encoder. For them the diagnosis is
direct and actionable. The contribution is a diagnosis, not a push to change
production systems that already use well-spread embedders. One geometry number,
like rogue-dimension dominance or the participation ratio, reveals whether
cosine is safe on a given encoder before it is trusted.

\section{Discussion}

This result changes the question. It is not which similarity metric is best in
general, since none wins everywhere, but what about the embedding space decides the
answer, and the answer is how much the variance piles into a few directions. Once
an encoder spreads its vectors out, cosine is already the best simple choice. On
crowded language-model vectors, the choice of metric really does matter. The
absolute gain there is modest, about $0.05$ in Spearman, but cosine starts from a
low base, so in relative terms it is sizable, near twenty percent on average and
almost sixty percent on GPT-2. It does not make these encoders match a well-spread
embedder, but it recovers a real part of what cosine was losing.

It is tempting to say the answer is just how the model was trained, since the
split lines up with the training objective. We resist that, because the metric
responds to the geometry the recipe leaves behind, not the recipe itself. When the
two come apart, the metric follows the geometry. E5-Mistral is trained with a
cosine objective but keeps a crowded space, and the alternatives help it.
Multilingual BERT is a plain language model but is well spread, and they do not.
Reading the geometry directly is more accurate and more useful than reading the
training label, since any encoder's geometry can be measured without knowing how
it was built.

This also explains why earlier reports disagreed. A paper that tries an
alternative on a plain language model will see it beat cosine. A paper that tries
the same metric on a well-spread embedder will see no gain. Both can be right at
the same time, because they sit on opposite sides of the geometric split. Our
rogue-dimension number says which side a given encoder is on, which turns a pile
of conflicting anecdotes into one rule.

The mechanism ties an old result to current practice. \citet{aggarwal2001surprising}
argued that $L_2$-based distances suffer in high dimensions, because a few large
coordinates take them over, and that $L_1$ and fractional norms hold up better.
Cosine is $L_2$-based, since it normalizes by the $L_2$ norm and scores with an
inner product. Our crowded encoders are exactly the case that argument
describes, with one coordinate carrying much of the variance, and the metrics
that win are the ones the argument points to. What we add is to show the
prediction holds on real embeddings and, by taking the crowded directions out,
to give evidence that those directions are the cause.

The practical message is narrow but useful. We are not telling anyone to drop
cosine in a deployed system, since those use well-spread encoders where cosine is
already best among parameter-free metrics. The point is to treat the encoder's
geometry as a precondition for trusting cosine. Someone using raw language-model
embeddings for similarity, perhaps to skip a fine-tuning step, can measure
rogue-dimension dominance first. If it is high, a rank-based or $L_1$-type metric
will do better, or the crowded directions can be removed so cosine recovers.

\section{Limitations}

Our encoders are mostly English and our datasets are short text, so we have not
checked cross-lingual matching or long documents, where the geometry might differ.
The gain is also small in absolute terms, about $0.05$ in Spearman, and although
that is a sizable relative gain against a low cosine baseline, it does not lift
these encoders to the level of a well-spread embedder.

Two smaller caveats remain. Our headline linear correlation of $0.95$ leans on
GPT-2, which sits far from the rest; dropping it gives $0.91$, which is why we
report the rank correlation of $0.86$ alongside it. And rogue-dimension dominance
is basis-dependent, though the principal version predicts almost as well and the
removal test works on principal directions, so the story does not hinge on the
basis.

\section{Conclusion}

We set out to learn when a simple metric beats cosine for comparing text
embeddings and what decides it. Across nineteen encoders and seven datasets we
found a clean split: cosine is the best parameter-free choice on encoders whose
vectors spread out, and it is clearly beaten by rank-based and $L_1$-type metrics
on encoders whose variance piles into a few directions. Model size, pooling, the
task, weak cosine, and vector length none explain it. The crowding of the space
does, measured with one number that predicts the gain at a rank correlation of
$0.86$ and a linear correlation of $0.95$, and removing the crowded directions
makes cosine recover.

The split tracks how the encoder was trained, but what the metric responds to is
the geometry, not the training, which turns a scattered set of claims into one
rule. For the commonly deployed fine-tuned embedders we tested, cosine is already
the right tool among parameter-free metrics. For raw language-model embeddings, the
geometry can be measured up front to say whether cosine is safe. The gain where it
applies is modest in absolute Spearman but sizable against a low baseline, around
twenty percent on average.

\subsubsection*{Broader Impact Statement}

This work studies how to compare text embeddings and does not introduce a new
model or dataset. The main effect on practice is a simple check that tells when
cosine similarity is reliable for a given encoder. We see little risk of harm
from such a check. If anything, it can catch quiet failures in systems that
compute similarity on encoders whose geometry makes cosine unreliable, which
pushes toward more careful evaluation rather than less.

\bibliography{main}

\begin{thebibliography}{51}
\providecommand{\natexlab}[1]{#1}
\providecommand{\url}[1]{\texttt{#1}}
\expandafter\ifx\csname urlstyle\endcsname\relax
  \providecommand{\doi}[1]{doi: #1}\else
  \providecommand{\doi}{doi: \begingroup \urlstyle{rm}\Url}\fi

\bibitem[Aggarwal et~al.(2001)Aggarwal, Hinneburg, and
  Keim]{aggarwal2001surprising}
Charu~C. Aggarwal, Alexander Hinneburg, and Daniel~A. Keim.
\newblock On the surprising behavior of distance metrics in high dimensional
  space.
\newblock \emph{Database Theory --- ICDT 2001}, pp.\  420--434, 2001.

\bibitem[Bajusz et~al.(2015)Bajusz, R{\'a}cz, and
  H{\'e}berger]{bajusz2015tanimoto}
D{\'a}vid Bajusz, Anita R{\'a}cz, and K{\'a}roly H{\'e}berger.
\newblock Why is {T}animoto index an appropriate choice for fingerprint-based
  similarity calculations?
\newblock \emph{Journal of Cheminformatics}, 7\penalty0 (1):\penalty0 20, 2015.

\bibitem[BehnamGhader et~al.(2024)BehnamGhader, Adlakha, Mosbach, Bahdanau,
  Chapados, and Reddy]{behnamghader2024llm2vec}
Parishad BehnamGhader, Vaibhav Adlakha, Marius Mosbach, Dzmitry Bahdanau,
  Nicolas Chapados, and Siva Reddy.
\newblock {LLM2Vec}: Large language models are secretly powerful text encoders.
\newblock In \emph{Conference on Language Modeling (COLM)}, 2024.

\bibitem[Bhattacharyya(1943)]{bhattacharyya1943measure}
Anil Bhattacharyya.
\newblock On a measure of divergence between two statistical populations
  defined by their probability distributions.
\newblock \emph{Bulletin of the Calcutta Mathematical Society}, 35:\penalty0
  99--109, 1943.

\bibitem[Biderman et~al.(2023)Biderman, Schoelkopf, Anthony, Bradley, O'Brien,
  Hallahan, Khan, Purohit, Prashanth, Raff, et~al.]{biderman2023pythia}
Stella Biderman, Hailey Schoelkopf, Quentin Anthony, Herbie Bradley, Kyle
  O'Brien, Eric Hallahan, Mohammad~Aflah Khan, Shivanshu Purohit, USVSN~Sai
  Prashanth, Edward Raff, et~al.
\newblock Pythia: A suite for analyzing large language models across training
  and scaling.
\newblock In \emph{Proceedings of the 40th International Conference on Machine
  Learning (ICML)}, pp.\  2397--2430, 2023.

\bibitem[Bowman et~al.(2015)Bowman, Angeli, Potts, and Manning]{bowman2015snli}
Samuel~R. Bowman, Gabor Angeli, Christopher Potts, and Christopher~D. Manning.
\newblock A large annotated corpus for learning natural language inference.
\newblock In \emph{Proceedings of the 2015 Conference on Empirical Methods in
  Natural Language Processing (EMNLP)}, pp.\  632--642, 2015.

\bibitem[Bray \& Curtis(1957)Bray and Curtis]{bray1957ordination}
J.~Roger Bray and John~T. Curtis.
\newblock An ordination of the upland forest communities of southern wisconsin.
\newblock \emph{Ecological Monographs}, 27\penalty0 (4):\penalty0 325--349,
  1957.

\bibitem[Cai et~al.(2021)Cai, Huang, Bian, and Church]{cai2021isotropy}
Xingyu Cai, Jiaji Huang, Yuchen Bian, and Kenneth Church.
\newblock Isotropy in the contextual embedding space: Clusters and manifolds.
\newblock In \emph{International Conference on Learning Representations
  (ICLR)}, 2021.

\bibitem[Cer et~al.(2017)Cer, Diab, Agirre, Lopez-Gazpio, and
  Specia]{cer2017semeval}
Daniel Cer, Mona Diab, Eneko Agirre, I{\~n}igo Lopez-Gazpio, and Lucia Specia.
\newblock Semeval-2017 task 1: Semantic textual similarity multilingual and
  crosslingual focused evaluation.
\newblock In \emph{Proceedings of the 11th International Workshop on Semantic
  Evaluation (SemEval-2017)}, pp.\  1--14, 2017.

\bibitem[Cha(2007)]{cha2007comprehensive}
Sung-Hyuk Cha.
\newblock Comprehensive survey on distance/similarity measures between
  probability density functions.
\newblock \emph{International Journal of Mathematical Models and Methods in
  Applied Sciences}, 1\penalty0 (4):\penalty0 300--307, 2007.

\bibitem[Clark et~al.(2020)Clark, Luong, Le, and Manning]{clark2020electra}
Kevin Clark, Minh-Thang Luong, Quoc~V. Le, and Christopher~D. Manning.
\newblock {ELECTRA}: Pre-training text encoders as discriminators rather than
  generators.
\newblock In \emph{International Conference on Learning Representations
  (ICLR)}, 2020.

\bibitem[Devlin et~al.(2019)Devlin, Chang, Lee, and Toutanova]{devlin2019bert}
Jacob Devlin, Ming-Wei Chang, Kenton Lee, and Kristina Toutanova.
\newblock {BERT}: Pre-training of deep bidirectional transformers for language
  understanding.
\newblock In \emph{Proceedings of the 2019 Conference of the North American
  Chapter of the Association for Computational Linguistics: Human Language
  Technologies (NAACL-HLT)}, pp.\  4171--4186, 2019.

\bibitem[Dice(1945)]{dice1945measures}
Lee~R. Dice.
\newblock Measures of the amount of ecologic association between species.
\newblock \emph{Ecology}, 26\penalty0 (3):\penalty0 297--302, 1945.

\bibitem[Ethayarajh(2019)]{ethayarajh2019contextual}
Kawin Ethayarajh.
\newblock How contextual are contextualized word representations? comparing the
  geometry of {BERT}, {ELMo}, and {GPT}-2 embeddings.
\newblock In \emph{Proceedings of the 2019 Conference on Empirical Methods in
  Natural Language Processing (EMNLP)}, pp.\  55--65, 2019.

\bibitem[Gao et~al.(2021)Gao, Yao, and Chen]{gao2021simcse}
Tianyu Gao, Xingcheng Yao, and Danqi Chen.
\newblock {SimCSE}: Simple contrastive learning of sentence embeddings.
\newblock In \emph{Proceedings of the 2021 Conference on Empirical Methods in
  Natural Language Processing (EMNLP)}, pp.\  6894--6910, 2021.

\bibitem[H{\"a}mmerl et~al.(2023)H{\"a}mmerl, Fastowski, Libovick{\'y}, and
  Fraser]{haemmerl2023anisotropy}
Katharina H{\"a}mmerl, Alina Fastowski, Jind{\v{r}}ich Libovick{\'y}, and
  Alexander Fraser.
\newblock Exploring anisotropy and outliers in multilingual language models for
  cross-lingual semantic sentence similarity.
\newblock In \emph{Findings of the Association for Computational Linguistics:
  ACL 2023}, pp.\  7023--7037, 2023.

\bibitem[Hellinger(1909)]{hellinger1909neue}
Ernst Hellinger.
\newblock Neue begr{\"u}ndung der theorie quadratischer formen von
  unendlichvielen ver{\"a}nderlichen.
\newblock \emph{Journal f{\"u}r die reine und angewandte Mathematik},
  136:\penalty0 210--271, 1909.

\bibitem[Huang et~al.(2021)Huang, Tang, Zhong, Lu, Shou, Gong, Jiang, and
  Duan]{huang2021whiteningbert}
Junjie Huang, Duyu Tang, Wanjun Zhong, Shuai Lu, Linjun Shou, Ming Gong, Daxin
  Jiang, and Nan Duan.
\newblock {WhiteningBERT}: An easy unsupervised sentence embedding approach.
\newblock \emph{Findings of the Association for Computational Linguistics:
  EMNLP 2021}, pp.\  238--244, 2021.

\bibitem[Jiang et~al.(2023)Jiang, Sablayrolles, Mensch, Bamford, Chaplot,
  de~las Casas, Bressand, Lengyel, Lample, Saulnier, et~al.]{jiang2023mistral}
Albert~Q. Jiang, Alexandre Sablayrolles, Arthur Mensch, Chris Bamford,
  Devendra~Singh Chaplot, Diego de~las Casas, Florian Bressand, Gianna Lengyel,
  Guillaume Lample, Lucile Saulnier, et~al.
\newblock Mistral 7{B}.
\newblock \emph{arXiv preprint arXiv:2310.06825}, 2023.

\bibitem[Kovaleva et~al.(2021)Kovaleva, Kulshreshtha, Rogers, and
  Rumshisky]{kovaleva2021bert}
Olga Kovaleva, Saurabh Kulshreshtha, Anna Rogers, and Anna Rumshisky.
\newblock {BERT} busters: Outlier dimensions that disrupt transformers.
\newblock In \emph{Findings of the Association for Computational Linguistics:
  ACL-IJCNLP 2021}, pp.\  3392--3405, 2021.

\bibitem[Lance \& Williams(1966)Lance and Williams]{lance1966computer}
Godfrey~N. Lance and William~T. Williams.
\newblock Computer programs for hierarchical polythetic classification
  (``similarity analyses'').
\newblock \emph{The Computer Journal}, 9\penalty0 (1):\penalty0 60--64, 1966.

\bibitem[Levy et~al.(2024)Levy, Shalom, and Chalamish]{levy2024guide}
Avivit Levy, B.~Riva Shalom, and Michal Chalamish.
\newblock A guide to similarity measures.
\newblock \emph{arXiv preprint arXiv:2408.07706}, 2024.

\bibitem[Li et~al.(2020)Li, Zhou, He, Wang, Yang, and Li]{li2020sentence}
Bohan Li, Hao Zhou, Junxian He, Mingxuan Wang, Yiming Yang, and Lei Li.
\newblock On the sentence embeddings from pre-trained language models.
\newblock In \emph{Proceedings of the 2020 Conference on Empirical Methods in
  Natural Language Processing (EMNLP)}, pp.\  9119--9130, 2020.

\bibitem[Lin(1991)]{lin1991divergence}
Jianhua Lin.
\newblock Divergence measures based on the shannon entropy.
\newblock \emph{IEEE Transactions on Information Theory}, 37\penalty0
  (1):\penalty0 145--151, 1991.

\bibitem[Liu et~al.(2019)Liu, Ott, Goyal, Du, Joshi, Chen, Levy, Lewis,
  Zettlemoyer, and Stoyanov]{liu2019roberta}
Yinhan Liu, Myle Ott, Naman Goyal, Jingfei Du, Mandar Joshi, Danqi Chen, Omer
  Levy, Mike Lewis, Luke Zettlemoyer, and Veselin Stoyanov.
\newblock {RoBERTa}: A robustly optimized {BERT} pretraining approach.
\newblock \emph{arXiv preprint arXiv:1907.11692}, 2019.

\bibitem[Marelli et~al.(2014)Marelli, Menini, Baroni, Bentivogli, Bernardi, and
  Zamparelli]{marelli2014sick}
Marco Marelli, Stefano Menini, Marco Baroni, Luisa Bentivogli, Raffaella
  Bernardi, and Roberto Zamparelli.
\newblock A {SICK} cure for the evaluation of compositional distributional
  semantic models.
\newblock In \emph{Proceedings of the Ninth International Conference on
  Language Resources and Evaluation (LREC)}, pp.\  216--223, 2014.

\bibitem[Meng et~al.(2024)Meng, Liu, Joty, Xiong, Zhou, and Yavuz]{meng2024sfr}
Rui Meng, Ye~Liu, Shafiq~Rayhan Joty, Caiming Xiong, Yingbo Zhou, and Semih
  Yavuz.
\newblock {SFR-Embedding-Mistral}: Enhance text retrieval with transfer
  learning.
\newblock \emph{Salesforce AI Research Blog}, 2024.

\bibitem[Mu \& Viswanath(2018)Mu and Viswanath]{mu2018allbutthetop}
Jiaqi Mu and Pramod Viswanath.
\newblock All-but-the-top: Simple and effective postprocessing for word
  representations.
\newblock In \emph{International Conference on Learning Representations
  (ICLR)}, 2018.

\bibitem[Muennighoff et~al.(2023)Muennighoff, Tazi, Magne, and
  Reimers]{muennighoff2023mteb}
Niklas Muennighoff, Nouamane Tazi, Lo{\"i}c Magne, and Nils Reimers.
\newblock {MTEB}: Massive text embedding benchmark.
\newblock In \emph{Proceedings of the 17th Conference of the European Chapter
  of the Association for Computational Linguistics}, pp.\  2014--2037, 2023.

\bibitem[Parupudi(2025)]{parupudi2025magnitude}
V.~S.~Raghu Parupudi.
\newblock Magnitude matters: a superior class of similarity metrics for
  holistic semantic understanding.
\newblock \emph{arXiv preprint arXiv:2509.19323}, 2025.

\bibitem[Puccetti et~al.(2022)Puccetti, Rogers, Drozd, and
  Dell'Orletta]{puccetti2022outlier}
Giovanni Puccetti, Anna Rogers, Aleksandr Drozd, and Felice Dell'Orletta.
\newblock Outlier dimensions that disrupt transformers are driven by frequency.
\newblock In \emph{Findings of the Association for Computational Linguistics:
  EMNLP 2022}, pp.\  1286--1304, 2022.

\bibitem[Radford et~al.(2019)Radford, Wu, Child, Luan, Amodei, and
  Sutskever]{radford2019gpt2}
Alec Radford, Jeffrey Wu, Rewon Child, David Luan, Dario Amodei, and Ilya
  Sutskever.
\newblock Language models are unsupervised multitask learners.
\newblock \emph{OpenAI Technical Report}, 2019.

\bibitem[Reimers \& Gurevych(2019)Reimers and Gurevych]{reimers2019sentence}
Nils Reimers and Iryna Gurevych.
\newblock Sentence-{BERT}: Sentence embeddings using {S}iamese {BERT}-networks.
\newblock In \emph{Proceedings of the 2019 Conference on Empirical Methods in
  Natural Language Processing (EMNLP)}, pp.\  3982--3992, 2019.

\bibitem[Rudman et~al.(2022)Rudman, Gillman, Rayne, and
  Eickhoff]{rudman2022isoscore}
William Rudman, Nate Gillman, Tyler Rayne, and Carsten Eickhoff.
\newblock {IsoScore}: Measuring the uniformity of embedding space utilization.
\newblock In \emph{Findings of the Association for Computational Linguistics:
  ACL 2022}, pp.\  3325--3339, 2022.

\bibitem[Song et~al.(2020)Song, Tan, Qin, Lu, and Liu]{song2020mpnet}
Kaitao Song, Xu~Tan, Tao Qin, Jianfeng Lu, and Tie-Yan Liu.
\newblock {MPNet}: Masked and permuted pre-training for language understanding.
\newblock In \emph{Advances in Neural Information Processing Systems
  (NeurIPS)}, volume~33, pp.\  16857--16867, 2020.

\bibitem[Spearman(1904)]{spearman1904proof}
Charles Spearman.
\newblock The proof and measurement of association between two things.
\newblock \emph{The American Journal of Psychology}, 15\penalty0 (1):\penalty0
  72--101, 1904.

\bibitem[Steck et~al.(2024)Steck, Ekanadham, and Kallus]{steck2024cosine}
Harald Steck, Chaitanya Ekanadham, and Nathan Kallus.
\newblock Is cosine-similarity of embeddings really about similarity?
\newblock In \emph{Companion Proceedings of the ACM Web Conference 2024 (WWW
  '24 Companion)}, pp.\  887--890, 2024.

\bibitem[Su et~al.(2021)Su, Cao, Liu, and Ou]{su2021whitening}
Jianlin Su, Jiarun Cao, Weijie Liu, and Yangyiwen Ou.
\newblock Whitening sentence representations for better semantics and faster
  retrieval.
\newblock \emph{arXiv preprint arXiv:2103.15316}, 2021.

\bibitem[Sun et~al.(2024)Sun, Chen, Kolter, and Liu]{sun2024massive}
Mingjie Sun, Xinlei Chen, J.~Zico Kolter, and Zhuang Liu.
\newblock Massive activations in large language models.
\newblock In \emph{Conference on Language Modeling (COLM)}, 2024.

\bibitem[Tan et~al.(2005)Tan, Steinbach, and Kumar]{tan2005introduction}
Pang-Ning Tan, Michael Steinbach, and Vipin Kumar.
\newblock \emph{Introduction to Data Mining}.
\newblock Addison-Wesley, 2005.

\bibitem[Team(2024)]{qwen2024report}
Qwen Team.
\newblock Qwen2.5 technical report.
\newblock \emph{arXiv preprint arXiv:2412.15115}, 2024.

\bibitem[Tessari et~al.(2025)Tessari, Yao, and Hogan]{tessari2025dimension}
Federico Tessari, Kunpeng Yao, and Neville Hogan.
\newblock Surpassing cosine similarity for multidimensional comparisons:
  Dimension insensitive euclidean metric.
\newblock \emph{arXiv preprint arXiv:2407.08623}, 2025.

\bibitem[Timkey \& van Schijndel(2021)Timkey and van Schijndel]{timkey2021bark}
William Timkey and Marten van Schijndel.
\newblock All bark and no bite: Rogue dimensions in transformer language models
  obscure representational quality.
\newblock In \emph{Proceedings of the 2021 Conference on Empirical Methods in
  Natural Language Processing (EMNLP)}, pp.\  4527--4546, 2021.

\bibitem[Wang et~al.(2019)Wang, Singh, Michael, Hill, Levy, and
  Bowman]{wang2019glue}
Alex Wang, Amanpreet Singh, Julian Michael, Felix Hill, Omer Levy, and
  Samuel~R. Bowman.
\newblock {GLUE}: A multi-task benchmark and analysis platform for natural
  language understanding.
\newblock In \emph{International Conference on Learning Representations
  (ICLR)}, 2019.

\bibitem[Wang et~al.(2022)Wang, Yang, Huang, Jiao, Yang, Jiang, Majumder, and
  Wei]{wang2022e5}
Liang Wang, Nan Yang, Xiaolong Huang, Binxing Jiao, Linjun Yang, Daxin Jiang,
  Rangan Majumder, and Furu Wei.
\newblock Text embeddings by weakly-supervised contrastive pre-training.
\newblock \emph{arXiv preprint arXiv:2212.03533}, 2022.

\bibitem[Wang et~al.(2024)Wang, Yang, Huang, Yang, Majumder, and
  Wei]{wang2024e5mistral}
Liang Wang, Nan Yang, Xiaolong Huang, Linjun Yang, Rangan Majumder, and Furu
  Wei.
\newblock Improving text embeddings with large language models.
\newblock In \emph{Proceedings of the 62nd Annual Meeting of the Association
  for Computational Linguistics (Volume 1: Long Papers)}, pp.\  11897--11916,
  2024.

\bibitem[Wang et~al.(2020)Wang, Wei, Dong, Bao, Yang, and Zhou]{wang2020minilm}
Wenhui Wang, Furu Wei, Li~Dong, Hangbo Bao, Nan Yang, and Ming Zhou.
\newblock {MiniLM}: Deep self-attention distillation for task-agnostic
  compression of pre-trained transformers.
\newblock In \emph{Advances in Neural Information Processing Systems
  (NeurIPS)}, volume~33, pp.\  5776--5788, 2020.

\bibitem[Williams et~al.(2018)Williams, Nangia, and
  Bowman]{williams2018multinli}
Adina Williams, Nikita Nangia, and Samuel~R. Bowman.
\newblock A broad-coverage challenge corpus for sentence understanding through
  inference.
\newblock In \emph{Proceedings of the 2018 Conference of the North American
  Chapter of the Association for Computational Linguistics: Human Language
  Technologies (NAACL-HLT)}, pp.\  1112--1122, 2018.

\bibitem[Xiao et~al.(2024)Xiao, Liu, Zhang, Muennighoff, Lian, and
  Nie]{xiao2023cpack}
Shitao Xiao, Zheng Liu, Peitian Zhang, Niklas Muennighoff, Defu Lian, and
  Jian-Yun Nie.
\newblock {C-Pack}: Packed resources for general chinese embeddings.
\newblock In \emph{Proceedings of the 47th International ACM SIGIR Conference
  on Research and Development in Information Retrieval}, 2024.

\bibitem[Zhang et~al.(2019)Zhang, Baldridge, and He]{zhang2019paws}
Yuan Zhang, Jason Baldridge, and Luheng He.
\newblock {PAWS}: Paraphrase adversaries from word scrambling.
\newblock In \emph{Proceedings of the 2019 Conference of the North American
  Chapter of the Association for Computational Linguistics: Human Language
  Technologies (NAACL-HLT)}, pp.\  1298--1308, 2019.

\bibitem[Zhou et~al.(2022)Zhou, Ethayarajh, Card, and
  Jurafsky]{zhou2022problems}
Kaitlyn Zhou, Kawin Ethayarajh, Dallas Card, and Dan Jurafsky.
\newblock Problems with cosine as a measure of embedding similarity for high
  frequency words.
\newblock In \emph{Proceedings of the 60th Annual Meeting of the Association
  for Computational Linguistics (Volume 2: Short Papers)}, pp.\  401--423,
  2022.

\end{thebibliography}
\bibliographystyle{tmlr}

\appendix
\section{Encoders and Geometry}

Table~\ref{tab:encoders} lists the nineteen encoders with their parameter count,
pooling method, and the geometry measures used in the analysis. The encoders are
sorted by rogue-dimension dominance, the share of variance held by the single
largest-variance dimension. The ordering tracks how the encoders were trained
but is not identical to it. Most contrastively trained embedders sit near the
bottom with low rogue-dimension dominance, and most base language models sit
higher, but E5-Mistral is a contrastively trained embedder that ranks among the
crowded ones, and multilingual BERT is a base model that ranks among the well
spread.

\begin{table}[t]
\caption{The nineteen encoders, sorted by rogue-dimension dominance. Params are
in millions. IsoScore is higher for more isotropic spaces. Part. ratio is the
participation ratio, an effective-dimensionality measure. RandCos is the mean
cosine between random pairs. Rogue is the variance share of the top dimension.}
\label{tab:encoders}
\begin{center}
\small
\begin{tabular}{lrlrrrr}
\toprule
Encoder & Params & Pool & IsoScore & Part.\ ratio & RandCos & Rogue \\
\midrule
GPT-2 & 124 & mean & 0.002 & 0.003 & 0.996 & 0.411 \\
Qwen2.5-7B & 7000 & last & 0.002 & 0.002 & 0.572 & 0.150 \\
Qwen2.5-1.5B & 1500 & last & 0.005 & 0.005 & 0.753 & 0.141 \\
ELECTRA-base & 110 & mean & 0.021 & 0.022 & 0.907 & 0.101 \\
RoBERTa-base & 125 & mean & 0.029 & 0.030 & 0.954 & 0.084 \\
Pythia-410M & 410 & last & 0.012 & 0.013 & 0.580 & 0.059 \\
Mistral-7B & 7000 & last & 0.004 & 0.004 & 0.466 & 0.032 \\
E5-Mistral-7B & 7000 & last & 0.008 & 0.008 & 0.406 & 0.026 \\
BERT-base & 110 & mean & 0.064 & 0.065 & 0.549 & 0.015 \\
mBERT-base & 178 & mean & 0.068 & 0.069 & 0.445 & 0.005 \\
SFR-Embedding-Mistral & 7000 & last & 0.028 & 0.028 & 0.415 & 0.005 \\
all-MiniLM-L6 & 22 & mean & 0.370 & 0.371 & 0.018 & 0.004 \\
paraphrase-mpnet & 110 & mean & 0.162 & 0.163 & 0.051 & 0.004 \\
all-MiniLM-L12 & 33 & mean & 0.393 & 0.394 & 0.019 & 0.004 \\
all-mpnet-base & 110 & mean & 0.222 & 0.222 & 0.016 & 0.003 \\
BGE-base & 110 & cls & 0.152 & 0.154 & 0.316 & 0.003 \\
multilingual-E5-large & 560 & mean & 0.116 & 0.116 & 0.707 & 0.002 \\
E5-large & 335 & mean & 0.118 & 0.119 & 0.696 & 0.002 \\
BGE-large & 335 & cls & 0.124 & 0.125 & 0.308 & 0.002 \\
\bottomrule
\end{tabular}
\end{center}
\end{table}

\section{Per-Metric Results}

Table~\ref{tab:bymetric} reports the mean Spearman gain of each metric over
cosine, split by whether the encoder is anisotropic. We call an encoder
anisotropic when its rogue-dimension dominance exceeds $0.01$. This puts nine
encoders on the crowded side and ten on the well-spread side. The split tracks
how the encoders were trained but is not identical to it. Most base models are
crowded and most contrastively trained embedders are well spread, but E5-Mistral
is a contrastively trained embedder that is crowded, and multilingual BERT is a
base model that is well spread, so the geometry split moves those two across the
training line. The metric pattern is clean either way. The rank-based and
$L_1$-type metrics gain on anisotropic encoders and do nothing on isotropic ones.
The angular family ties cosine in both groups. The information-geometry metrics
and the raw dot product lose, and they lose most on the anisotropic encoders.

\begin{table}[t]
\caption{Mean Spearman gain over cosine, averaged across encoders and datasets,
split by encoder geometry. Positive means the metric beats cosine. The rank-based
and $L_1$-type metrics are the only consistent winners, and only on anisotropic
encoders.}
\label{tab:bymetric}
\begin{center}
\small
\begin{tabular}{lrr}
\toprule
Metric & Anisotropic gain & Isotropic gain \\
\midrule
Cosine (reference) & $0.000$ & $0.000$ \\
\midrule
Spearman-vector & $+0.053$ & $-0.000$ \\
Canberra & $+0.051$ & $-0.004$ \\
Bray-Curtis & $+0.039$ & $-0.001$ \\
Fractional $L_{0.5}$ & $+0.034$ & $-0.001$ \\
Manhattan ($L_1$) & $+0.028$ & $-0.001$ \\
\midrule
Angular & $+0.000$ & $-0.000$ \\
Correlation & $+0.000$ & $+0.000$ \\
Tanimoto & $-0.009$ & $-0.000$ \\
Overlap & $-0.009$ & $-0.000$ \\
Hyperbolic tangent & $-0.009$ & $-0.000$ \\
Dice & $-0.009$ & $-0.000$ \\
\midrule
Euclidean & $-0.006$ & $-0.000$ \\
Gaussian kernel & $-0.006$ & $-0.000$ \\
Chebyshev ($L_\infty$) & $-0.115$ & $-0.042$ \\
Jensen-Shannon & $-0.143$ & $-0.000$ \\
Hellinger & $-0.144$ & $-0.000$ \\
Bhattacharyya & $-0.143$ & $-0.000$ \\
Dot product & $-0.152$ & $-0.019$ \\
\bottomrule
\end{tabular}
\end{center}
\end{table}

\section{Datasets and Splits}

We use seven datasets. For semantic textual similarity we use STS-B from the
GLUE benchmark on the validation split, SICK-R on the test split, and STS16 on
the test split, all with graded scores normalized to the unit interval. For
paraphrase detection we use Quora question pairs and PAWS on the labeled-final
test split, both with binary labels. For natural language inference we use SNLI
on the test split and MultiNLI on the matched validation split, mapping
entailment to one and the other labels to zero, and dropping examples with no
gold label. All datasets are loaded from their standard public sources.

\section{Per-Dataset Gains}

Table~\ref{tab:bydataset} breaks the headline result down to each of the seven
datasets, so the gain can be read off one dataset at a time rather than only as a
task-group average. For each dataset we report the mean cosine Spearman and the
mean gain of the best alternative metric over cosine, split by encoder geometry.
The pattern is the same on every dataset. On the crowded encoders the best
alternative beats cosine on all seven, by between $0.030$ and $0.087$, and on the
well-spread encoders the gain is at most $0.002$ everywhere. The effect is not
carried by any single dataset, and it does not depend on the task type. The gain
is largest where cosine is weakest, which is why the two inference datasets, where
cosine sits lowest, show some of the biggest gains. The rightmost column makes
this concrete in relative terms: because cosine starts low on the crowded
encoders, the same absolute gains run from eight percent on SICK-R up to seventy
percent on SNLI.

\begin{table}[t]
\caption{Mean cosine Spearman and mean best-alternative gain over cosine, per
dataset, split by encoder geometry. The best alternative is the best of the five
winning metrics (Spearman-vector, Canberra, Bray-Curtis, fractional $L_{0.5}$,
Manhattan) on each encoder-dataset cell. Positive gain means the alternative beats
cosine. Rel.\ is the gain as a percentage of the cosine baseline on the crowded
encoders. The gain is positive on every dataset for crowded encoders and near zero
for well-spread ones.}
\label{tab:bydataset}
\begin{center}
\small
\begin{tabular}{llrrrrr}
\toprule
& & \multicolumn{3}{c}{Crowded (9 enc.)} & \multicolumn{2}{c}{Well spread (10 enc.)} \\
\cmidrule(lr){3-5}\cmidrule(lr){6-7}
Task & Dataset & Cosine & Gain & Rel. & Cosine & Gain \\
\midrule
Similarity & STS-B & $0.479$ & $+0.087$ & $+18\%$ & $0.857$ & $+0.000$ \\
Similarity & SICK-R & $0.500$ & $+0.042$ & $+8\%$ & $0.774$ & $+0.002$ \\
Similarity & STS16 & $0.510$ & $+0.064$ & $+13\%$ & $0.810$ & $+0.001$ \\
\midrule
Paraphrase & Quora & $0.396$ & $+0.043$ & $+11\%$ & $0.593$ & $+0.000$ \\
Paraphrase & PAWS & $0.142$ & $+0.030$ & $+21\%$ & $0.236$ & $+0.000$ \\
\midrule
Inference & SNLI & $0.089$ & $+0.062$ & $+70\%$ & $0.458$ & $+0.002$ \\
Inference & MultiNLI & $0.153$ & $+0.053$ & $+35\%$ & $0.489$ & $-0.001$ \\
\bottomrule
\end{tabular}
\end{center}
\end{table}

\section{Chunked Computation}

To bound memory on the seven-billion-parameter encoders, we compute metric scores
in chunks of at most two thousand sentence pairs. Each chunk stores only the raw
metric score vectors and the gold labels. We then concatenate the chunks back
into full-length score vectors and compute every statistic once on the full data.
We verified that this matches the full-data computation exactly. For example,
the raw cosine Spearman for one encoder-dataset pair is $0.40458146$ under both
the chunked and the direct computation.

\end{document}